\documentclass[letterpaper, 10 pt, conference]{ieeeconf}  

\IEEEoverridecommandlockouts                              

\usepackage{amsmath}
\usepackage{amssymb}   
\usepackage{graphicx}
\usepackage{caption}
\usepackage{subcaption}
\usepackage{float}
\usepackage{booktabs} 
\usepackage{multirow}
\usepackage{makecell}
\usepackage{array}
\usepackage{algorithm}
\usepackage{algpseudocode} 
\usepackage{amsmath} 
\usepackage{amsmath} 
\usepackage{algorithm}
\usepackage{algpseudocode}

\usepackage{graphicx}
\usepackage{booktabs}
\usepackage{multirow}
\usepackage{adjustbox} 
\usepackage{colortbl}  
\usepackage{xcolor}
\definecolor{tabblue}{rgb}{0.88, 0.92, 0.98} 

\title{\LARGE \bf
Master–Micro Residual Correction with Adaptive Tactile Fusion and Force-Mixed Control for Contact-Rich Manipulation
}

\author{{Xingting Li}$^{1}$$^{*}$,
{Yifan Xie}$^{1}$$^{*}$,
{Han Liu}$^{1}$$^{*}$,
{Wei Hou}$^{1}$,
{Guangyu Chen}$^{2}$,
{Shoujie Li}$^{3}$,
{Wenbo Ding}$^{1}$$^{\dagger}$
\thanks{{$*$}These authors contributed equally to this
work.}
\thanks{{$^1$}Shenzhen Ubiquitous Data Enabling Key Lab, Shenzhen International Graduate School, Tsinghua University, Shenzhen 518055, China.}
\thanks{{$^2$}School of Electronics and Communication Engineering, Sun Yat-sen University, Shenzhen 518107, China.}
\thanks{{$^3$}School of Mechanical and Aerospace Engineering, Nanyang Technological University, Singapore 639956, Singapore.}
\thanks{{$\dagger$}Corresponding author: {ding.wenbo@sz.tsinghua.edu.cn}.}
}

\begin{document}

\maketitle
\thispagestyle{empty}
\pagestyle{empty}

\begin{abstract}

Robotic contact-rich and fine-grained manipulation remains a significant challenge due to complex interaction dynamics and the competing requirements of multi-timescale control. While current visual imitation learning methods excel at long-horizon planning, they often fail to perceive critical interaction cues like friction variations or incipient slip, and struggle to balance global task coherence with local reactive feedback. To address these challenges, we propose M\textsuperscript{2}-ResiPolicy, a novel Master-Micro residual control architecture that synergizes high-level action guidance with low-level correction. The framework consists of a Master-Guidance Policy (MGP) operating at 10 Hz, which generates temporally consistent action chunks via a diffusion-based backbone and employs a tactile-intensity-driven adaptive fusion mechanism to dynamically modulate perceptual weights between vision and touch. Simultaneously, a high-frequency (60 Hz) Micro-Residual Corrector (MRC) utilizes a lightweight GRU to provide real-time action compensation based on TCP wrench feedback. This policy is further integrated with a force-mixed PBIC execution layer, effectively regulating contact forces to ensure interaction safety. Experiments across several demanding tasks including fragile object grasping and precision insertion, demonstrate that M\textsuperscript{2}-ResiPolicy significantly outperforms standard Diffusion Policy (DP) and state-of-the-art Reactive Diffusion Policy (RDP), achieving a 93\% damage-free success rate in chip grasping and superior force regulation stability.

\end{abstract}

\begin{keywords} 
	Contact-Rich manipulation,
	Visuotactile fusion, 
    Residual policy learning.
\end{keywords}

\section{INTRODUCTION}

Robotic contact-rich and fine-grained manipulation~\cite{he2025foar,xue2025reactive,jiang2025gelfusion,li2025adaptive}, such as peeling vegetables or handling fragile objects, remains a significant challenge due to complex and partially observable interaction dynamics. 
Humans achieve such dexterity by naturally integrating vision and touch~\cite{jones2025beyond}: vision provides global scene context and task understanding, while tactile feedback supplies localized, contact-specific information essential for fine-grained control. 
Despite substantial progress in long-horizon visual imitation learning (IL)~\cite{yu2025forcevla,zhao2025cot}, vision-only policies often fail to capture frictional variations or incipient slip, limiting precision and stability in unstructured environments.

To overcome these limitations, prior work explores various visuo-tactile fusion strategies. Reactive Diffusion Policy~\cite{xue2025reactive} adopts a slow--fast hierarchical control paradigm that combines diffusion-based action chunking with high-frequency tactile feedback, while ViTacFormer~\cite{heng2025vitacformer} learns unified visuo--tactile representations via cross-attention and autoregressive tactile prediction for robust long-horizon manipulation. 
Nevertheless, existing hierarchical or temporally structured approaches still face a fundamental challenge: reconciling task-level temporal coherence with rapid, contact-sensitive force feedback. 
In particular, how perceptual focus should be dynamically modulated across contact phases and how physical safety can be systematically enforced at the execution layer remain largely unexplored.

\begin{figure}[!t]
    \centering
    \includegraphics[width=1.0\linewidth]{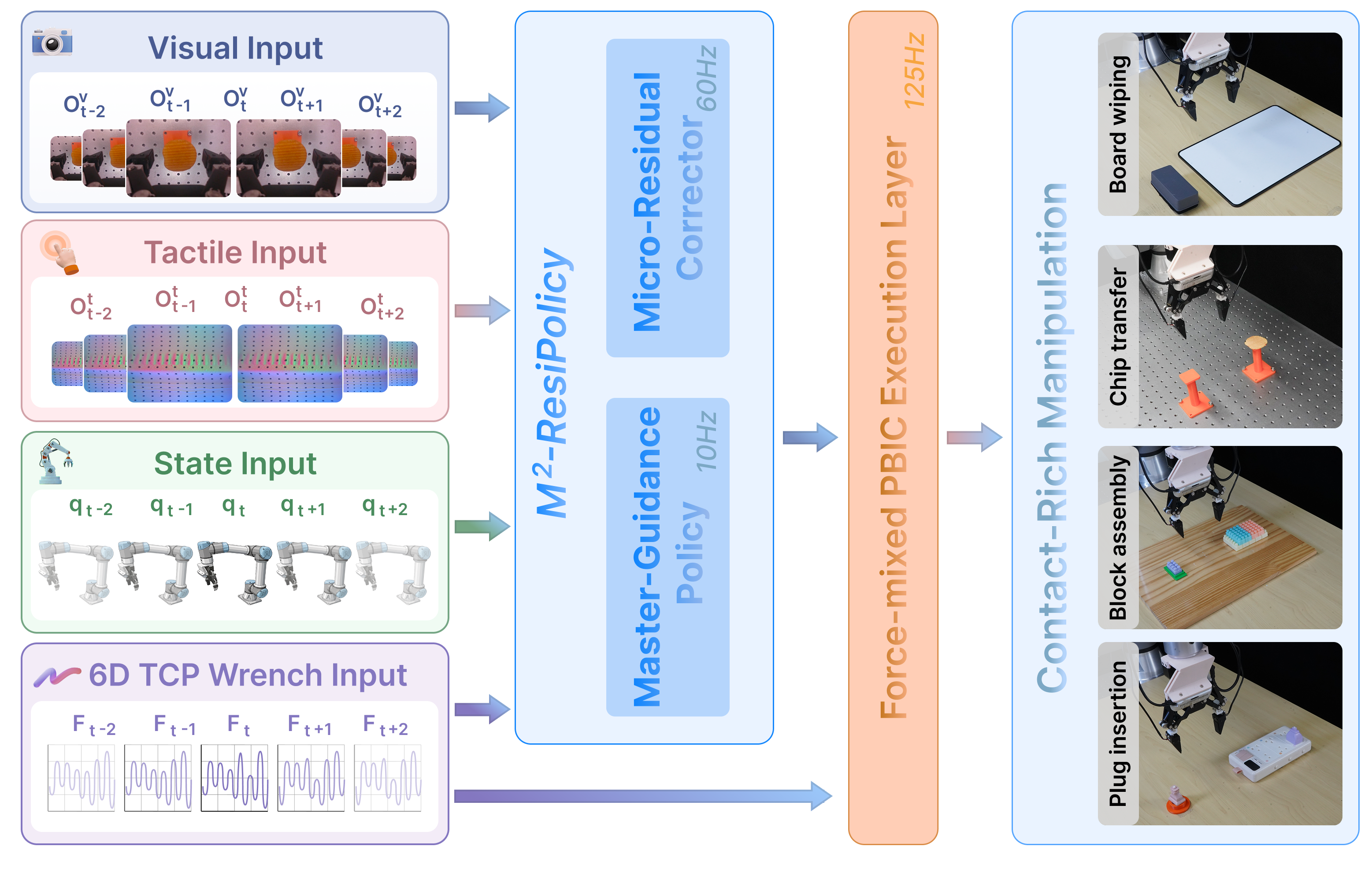}
    \caption{\textbf{M\textsuperscript{2}-ResiPolicy Framework.} Our dual-stream architecture decouples global planning (MGP) from high-frequency local correction (MRC), unified by a force-mixed PBIC execution layer.}
    \label{fig:teaser}
\end{figure}

To address these challenges, we propose \textbf{M\textsuperscript{2}-ResiPolicy}, a Master--Micro residual control framework for contact-rich manipulation that couples multi-timescale policy learning with compliant execution. At the high level, a \textit{Master-Guidance Policy} (MGP) uses confidence-gated visuotactile fusion to generate temporally coherent action chunks from multimodal observations. At the low level, a high-frequency \textit{Micro-Residual Corrector} (MRC) predicts residual compensation from TCP wrench feedback for rapid local refinement under transient contact. During execution, the policy output is passed through a force-mixed PBIC execution layer, which converts motion references into compliant pose commands to guarantee physical safety during contact-rich interactions.
\begin{itemize}
    \item[(1)] A confidence-gated visuotactile cross-attention fusion mechanism that adaptively modulates modality contributions across contact phases;
    \item[(2)] A high-frequency micro-residual correction module that uses TCP force/torque feedback for rapid local refinement under transient contact; and
    \item[(3)] A force-mixed PBIC execution layer that establishes a structured compliant action space, effectively regulating interaction forces to ensure physical safety and execution stability under contact.
\end{itemize}


\section{RELATED WORK}
\subsection{Learning Perception for Contact: Tactile and Force Representations}

Modeling physical interaction requires inferring latent contact states, such as contact location, friction, and incipient slip, that are not directly observable from vision alone. This motivates multimodal sensing, where vision provides global context, force/torque captures interaction intensity, and tactile sensing provides localized contact structure. Existing work can be broadly grouped by sensing modality and its role in policy learning.

\textbf{Vision-based tactile sensing} leverages image-form tactile signals to support adaptive multimodal fusion \cite{li2025adaptive}, robot-free visuo-tactile representation learning \cite{liu2025vitamin}, and dual-rate closed-loop control \cite{xue2025reactive}, where slow and fast loops couple task-level progression with rapid local adjustment.
Complementary to control-oriented methods, universal visuo-tactile video understanding aims to learn transferable cross-modal representations from large-scale tactile video data \cite{xie2025universal}.

In parallel, \textbf{proprioceptive force/torque (F/T) sensing} offers a low-dimensional yet highly responsive signal for detecting contact transitions and interaction directions. 
Force has been extensively incorporated into imitation learning, either as an explicit control target \cite{liu2025forcemimic, chen2025dexforce}, 
or through training and fusion schemes that encourage force-aware behaviors, including curriculum-based force attending \cite{liu2025factr} 
and phase-aware force utilization guided by contact prediction \cite{he2025foar}.

More recent efforts integrate force into conditional generative policies, including diffusion-based force-domain control \cite{wu2025tacdiffusion}, vision--force-guided compliant manipulation \cite{kang2025robotic}, and diffusion-guided compliance learning \cite{hou2025adaptive}.
Despite these advances, most existing frameworks either operate on a single modality or adopt static fusion schemes that do not dynamically adjust perceptual emphasis across different contact phases. 
Although phase-aware designs have been explored \cite{he2025foar}, achieving smooth and safe transitions from pre-contact to sustained-contact interaction remains an open challenge.

\begin{figure*}[!t]
    \centering
    \includegraphics[width=0.95\textwidth]{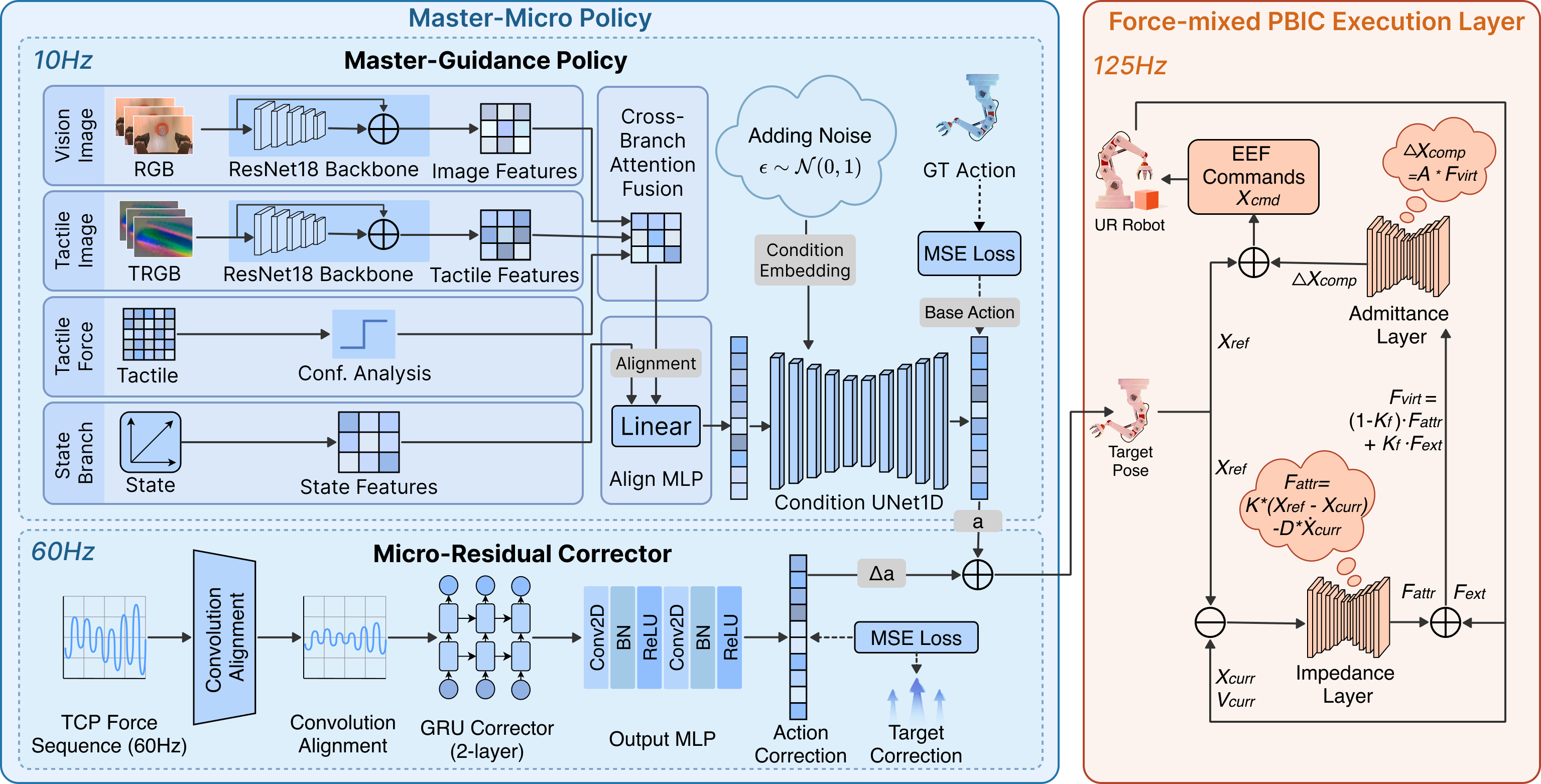}
    \caption{\textbf{Overview of the proposed M\textsuperscript{2}-ResiPolicy framework.}
    \emph{Left:} The 10~Hz Master-Guidance Policy encodes multimodal observations and generates temporally coherent action chunks via confidence-gated cross-attention fusion, while the 60~Hz Micro-Residual Corrector predicts residual compensation from aligned TCP wrench feedback.
    \emph{Right:} The reference pose $x_{\mathrm{ref}}$ is executed through the force-mixed PBIC layer at 125~Hz to produce a compliant pose command $x_{\mathrm{cmd}}$ for stable contact interaction.}
    \label{fig:slow_fast_framework}
\end{figure*}

\subsection{Multimodal Policy Learning for Contact-Rich Manipulation}

Recent work increasingly adopts \textbf{end-to-end multimodal policy learning}, directly mapping vision, tactile, and force observations to control commands. 
Early approaches combine vision and touch in deep reinforcement learning \cite{hansen2022visuotactile_rl}, while more recent work leverages large-scale visuotactile datasets to improve generalization \cite{yang2022touch_and_go}. 
Other methods decompose manipulation into reaching and local contact-centric interaction stages, enabling reusable tactile skills \cite{zhao2025touch_begins}, including extensions to simulation fine-tuning and bimanual assembly \cite{huang2025vt_refine}. 
However, most of these methods remain pose-centric at the execution level, issuing position or trajectory commands without explicitly modeling interaction compliance, which can compromise safety under contact. 
Recent work therefore begins to integrate impedance- or compliance-aware controllers into low-level execution for safer contact-rich manipulation \cite{hou2025adaptive, ge2025filic}.

Beyond deterministic regression, \textbf{conditional generative policies} model action distributions conditioned on perceptual context. 
Flow- and diffusion-based methods enable long-horizon, multi-modal action generation, including consistency-based flow matching \cite{yan2025maniflow} and tactile-conditioned, force-aware diffusion policies \cite{helmut2025tactile_conditioned_diffusion}. 
A key challenge remains the mismatch between temporal abstraction and control bandwidth: long-horizon policies smooth rapid contact variations, while highly reactive controllers sacrifice long-term coherence, motivating two-time-scale or hierarchical designs that decouple slow task evolution from fast, contact-driven adjustment.

\section{Method}\label{sec:method}

Contact-rich manipulation imposes competing multi-timescale requirements: the policy must maintain temporally consistent motion over long horizons, while remaining sufficiently reactive to rapid contact variations such as impacts and abrupt friction changes. To address this challenge, we propose \textbf{M\textsuperscript{2}-ResiPolicy}, a Master--Micro framework illustrated in Fig.~\ref{fig:slow_fast_framework}. At the high level, the \textit{Master-Guidance Policy} (MGP) predicts an $H$-step, 7D end-effector action chunk from multimodal observations at 10~Hz, providing a smooth global motion prior. At the low level, the \textit{Micro-Residual Corrector} (MRC) processes 6D TCP wrench feedback at 60~Hz and predicts temporally aligned residual corrections to refine execution under fast contact dynamics.

During deployment, the policy output is interpreted as a motion-reference pose $x_{\mathrm{ref}}$. To improve robustness under modeling errors and out-of-distribution contacts, we execute this reference through a position-based hybrid impedance controller, which combines an impedance-derived force intention with the measured external wrench to generate a compliant pose command $x_{\mathrm{cmd}} = x_{\mathrm{ref}} + \Delta x_{\mathrm{comp}}$. As an Attractor-Anchored Hybrid Compliance Layer, it preserves accurate free-space tracking while actively regulating forces, thereby ensuring physical safety in contact-rich interactions.

\subsection{Data Collection and System Setup}

As shown in Fig.~2, our data-collection platform consists of a UR7e manipulator teleoperated via a GELLO device~\cite{wu2024gello}, which maps the operator's end-effector pose intent to robot motion. Visual observations are provided by two RealSense D435 cameras, including a third-person view for global context and a wrist-mounted view for local details. Tactile observations are captured by two fingertip-mounted XENSE-G1 visuotactile sensors. Instead of using raw tactile images, we adopt frame-differenced tactile representations to improve signal consistency and stability. Force sensing is provided by the UR7e's built-in 6D TCP wrench, which captures fast interaction events such as contact onset, slip, and impact. All modalities are timestamped and segmented into episodes.

To support cross-rate learning, we record multimodal observations and end-effector action labels at 10~Hz, while logging TCP wrench measurements at 60~Hz. During offline preprocessing, the high-rate wrench stream is timestamp-aligned with the action steps so that each step is associated with the corresponding force/torque information while preserving temporal consistency. Demonstrations are collected under \textit{Position-Based Hybrid Impedance Control}, with task-specific stiffness, damping, and force-mixing gains tuned according to contact intensity and desired compliance. As a result, the dataset captures not only geometric motion intent but also stable interaction regularities induced by compliant regulation, enabling the learned policy to reproduce safe and compliant behaviors when deployed through the same execution layer.

\subsection{Confidence-Gated Cross-Attention Fusion in the Master-Guidance Policy}

\textit{Master-Guidance Policy (MGP)} runs at 10~Hz and predicts temporally coherent end-effector action chunks from multimodal observations, providing a smooth global motion prior for downstream execution. Let $\mathbf{o}_t$ denote the observation at time $t$, including third-person and wrist-view RGB images, two fingertip XENSE-G1 visuotactile observations, the robot proprioceptive state, and a synchronized 6D fingertip force/torque signal. The policy outputs an $H$-step 7D action chunk $\mathbf{a}_{t:t+H-1}\in\mathbb{R}^{H\times 7}$, modeled in a normalized action space during training.

As shown in Fig.~\ref{fig:slow_fast_framework} (top left), the RGB streams are encoded into visual tokens and the fingertip visuotactile streams into tactile tokens, while the proprioceptive state is retained as a low-dimensional auxiliary feature. Instead of applying generic self-attention over all modalities, MGP adopts a confidence-gated cross-attention module, where visual tokens act as the primary geometric anchor and tactile cues are incorporated according to the current contact state.

Let $\mathbf{F}_{v,t}\in\mathbb{R}^{N_r\times D}$ and $\mathbf{F}_{\mathrm{tac},t}\in\mathbb{R}^{N_t\times D}$ denote the visual and tactile tokens at time $t$, and let $\mathbf{f}_t\in\mathbb{R}^{6}$ be the corresponding fingertip force/torque signal. A binary contact confidence is first derived as
\begin{equation}
c_t=\mathbb{I}\!\left(\|\mathbf{f}_t\|_2>\tau\right),
\end{equation}
where $\tau$ is a small force threshold. The visual and tactile tokens are then concatenated into a joint key-value pool, while the visual tokens are used as queries~\cite{vaswani2017attention}. Specifically, with $\mathbf{Q}_t=\mathbf{W}_Q\mathbf{F}_{v,t}$ and $\mathbf{K}_t,\mathbf{V}_t$ obtained from the concatenated tokens, the masked cross-attention is computed as
\begin{equation}
\hat{\mathbf{S}}_t=
\left(\frac{\mathbf{Q}_t\mathbf{K}_t^\top}{\sqrt{d_h}}\right)\odot \mathbf{M}_t,
\end{equation}
where $\mathbf{M}_t$ preserves the visual logits and gates the tactile logits according to $c_t$. The fused feature is then obtained by
\begin{equation}
\mathbf{F}^{\mathrm{fused}}_t=
\mathbf{F}_{v,t}+\mathbf{W}_O\,\mathrm{MHA}(\mathbf{Q}_t,\mathbf{K}_t,\mathbf{V}_t).
\end{equation}
This design preserves vision as the global reference while allowing tactile cues to contribute under contact. The fused features are finally flattened, concatenated with the low-dimensional proprioceptive features, and mapped by a linear alignment layer to obtain the conditioning context $\mathbf{c}_t$.

The diffusion denoiser is trained with a conditional noise-prediction objective~\cite{chi2025diffusion}. Let $\tilde{\mathbf{a}}$ denote the normalized expert action chunk. We sample a diffusion step $k$ and Gaussian noise $\boldsymbol{\epsilon}\sim\mathcal{N}(\mathbf{0},\mathbf{I})$, and construct the noisy action as $\tilde{\mathbf{a}}_k=\sqrt{\bar{\alpha}_k}\tilde{\mathbf{a}}+\sqrt{1-\bar{\alpha}_k}\boldsymbol{\epsilon}$. The training objective is $\mathcal{L}_{\mathrm{MGP}}=\left\|\pi_{\mathrm{MGP}}(\tilde{\mathbf{a}}_k,k,\mathbf{c}_t)-\boldsymbol{\epsilon}\right\|_2^2$. At inference time, the same conditioning pathway guides iterative denoising to generate an $H$-step action chunk for subsequent compliant execution.

\begin{figure}[t]
    \centering
    \includegraphics[width=0.95\linewidth]{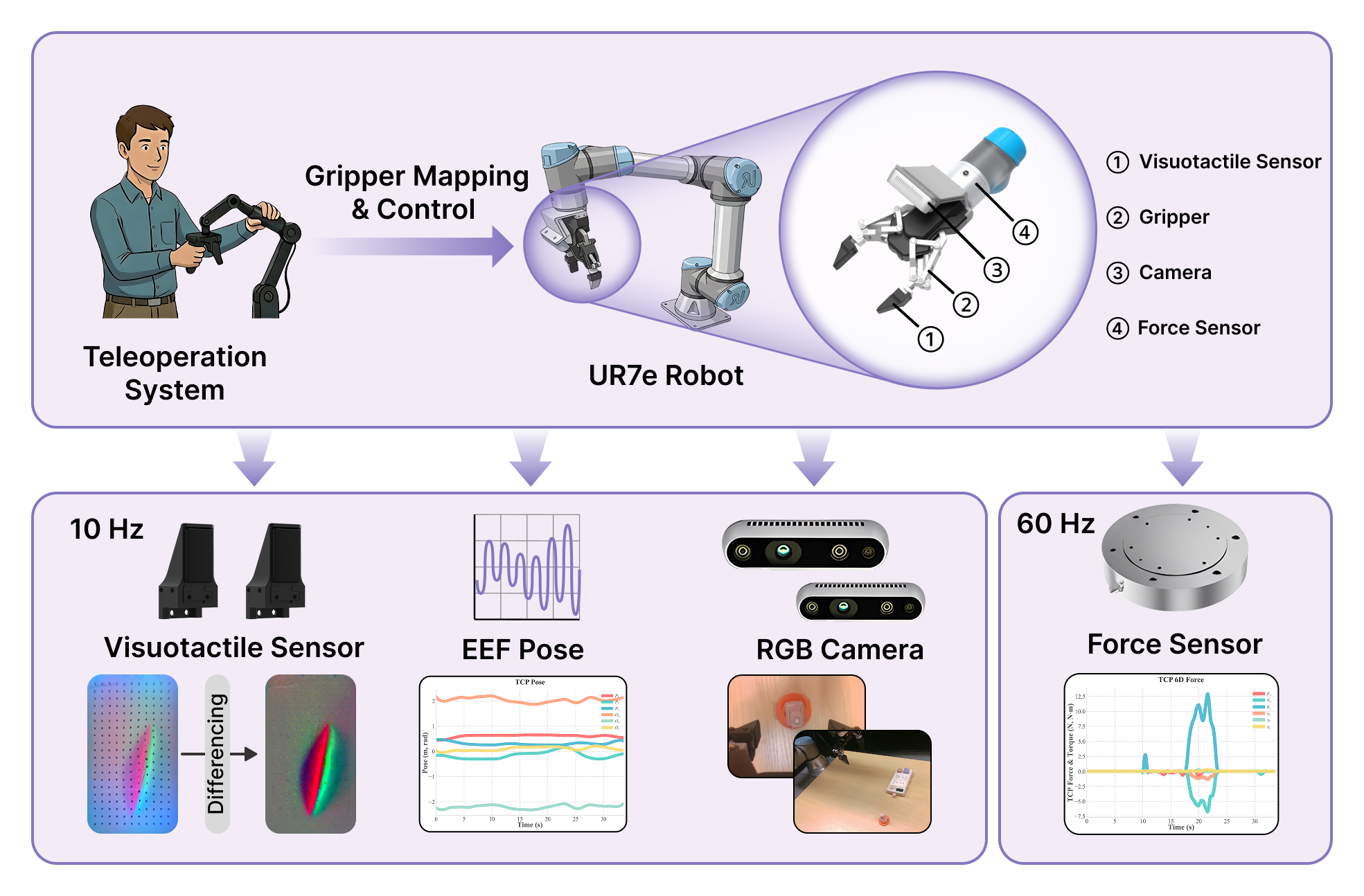}
    \caption{\textbf{System setup and data streams.} A teleoperated UR7e platform collects visuotactile, RGB, pose, and TCP wrench data. Visuotactile signals, RGB, and end-effector pose are sampled at 10~Hz, while TCP wrench is sampled at 60~Hz.}
    \label{fig:setup}
\end{figure}

\subsection{Reactive Micro-Residual Correction and Force-Mixed Compliant Execution}

The \textit{Micro-Residual Corrector (MRC)} is designed to capture high-frequency contact dynamics and suppress local execution deviations, such as those induced by impacts, abrupt friction transitions, and incipient slip. MRC operates in conjunction with the master action chunk predicted by the \textit{Master-Guidance Policy (MGP)}. The system continuously acquires 6D TCP force/torque (wrench) measurements at 60~Hz and compresses this high-rate data stream into a force-feature sequence aligned with the master action steps through strided temporal convolution. This ensures that each action step is paired with a compact summary of recent contact feedback. On top of these step-aligned features, MRC utilizes a lightweight GRU to model temporal dependencies, accumulating short-horizon contact history in its hidden state and predicting residual corrections aligned with the master action. This recurrent approach enables high-frequency contact information to be injected into residual predictions. To ensure consistency between training and inference, residuals are defined in the normalized action space, and the supervised target is constructed relative to a frozen MGP baseline: $\Delta\tilde{\mathbf{a}}^{*}=\tilde{\mathbf{a}}^{\mathrm{exp}}-\tilde{\mathbf{a}}^{\mathrm{MGP}}$. This formulation encourages MRC to learn local corrections around the master baseline, thereby improving the stability and transferability of contact-rich closed-loop execution.

During execution, a force-mixed PBIC execution layer (Algorithm 1) is integrated to formulate a compliant interface for the policy. Diverging from traditional wrench-only admittance control that strictly adheres to passivity, PBIC extracts a virtual attractor wrench from the tracking error and blends it with external wrench feedback, computing a compliance offset to yield $x_{\mathrm{cmd}} = x_{\mathrm{ref}} + \Delta x_{\mathrm{comp}}$ at 125~Hz. While introducing error-based forces into the execution loop might theoretically raise concerns about high-frequency oscillations, we strictly bound the admittance gain to a minimal magnitude. This ensures the mixed term acts practically as a controlled micro-feedforward compensation to overcome joint stiction, rather than a divergent positive feedback. Furthermore, the high-frequency Micro-Residual Corrector (MRC) naturally adapts to and compensates for these underlying dynamics during closed-loop learning. Consequently, this attractor-driven force mixing actively regulates contact forces, acting as a reliable safety buffer to ensure physical safety and execution stability in contact-rich interactions.

\begin{algorithm}[t]
\caption{Force-Mixed PBIC Execution}
\label{alg:hybrid_impedance}
\begin{algorithmic}[1]
\Require $\mathbf{x}_{\text{ref}}$ \Comment{Reference pose from high-level policy}
\Require $\mathbf{x}_{\text{curr}}, \dot{\mathbf{x}}_{\text{curr}}$ \Comment{Current robot state}
\Require $\mathbf{F}_{\text{ext}}$ \Comment{Measured external wrench}
\Require $\mathbf{K}, \mathbf{D}, \mathbf{A}, k_f$ \Comment{Control parameters}

\State \textit{Compute virtual attractor wrench from tracking error}
\State $\mathbf{F}_{\text{attr}} \leftarrow \mathbf{K}(\mathbf{x}_{\text{ref}} - \mathbf{x}_{\text{curr}}) - \mathbf{D}\dot{\mathbf{x}}_{\text{curr}}$

\State \textit{Formulate force-mixed virtual wrench}
\State $\mathbf{F}_{\text{virt}} \leftarrow (1 - k_f)\mathbf{F}_{\text{attr}} + k_f \mathbf{F}_{\text{ext}}$

\State \textit{Map to compliance offset via micro-admittance interface}
\State $\Delta \mathbf{x}_{\text{comp}} \leftarrow \mathbf{A} \mathbf{F}_{\text{virt}}$

\State \textit{Synthesize compliant action target}
\State $\mathbf{x}_{\text{cmd}} \leftarrow \mathbf{x}_{\text{ref}} + \Delta \mathbf{x}_{\text{comp}}$

\State \Return $\mathbf{x}_{\text{cmd}}$ \Comment{To position servo}
\end{algorithmic}
\end{algorithm}

\subsection{Staged Training Strategy}

We adopt a staged training scheme to decouple master action generation from micro-level residual correction. The master policy is first trained to produce reliable action chunks and define the action-normalization mapping. The master branch is then frozen, and the GRU-based residual module is trained to predict local corrections from the 6D end-effector force/torque signal around this fixed baseline. This training strategy reduces optimization interference and improves rapid contact adaptation while preserving long-horizon motion consistency.


\begin{figure*}[!ht]
    \centering
    \includegraphics[width=0.95\linewidth]{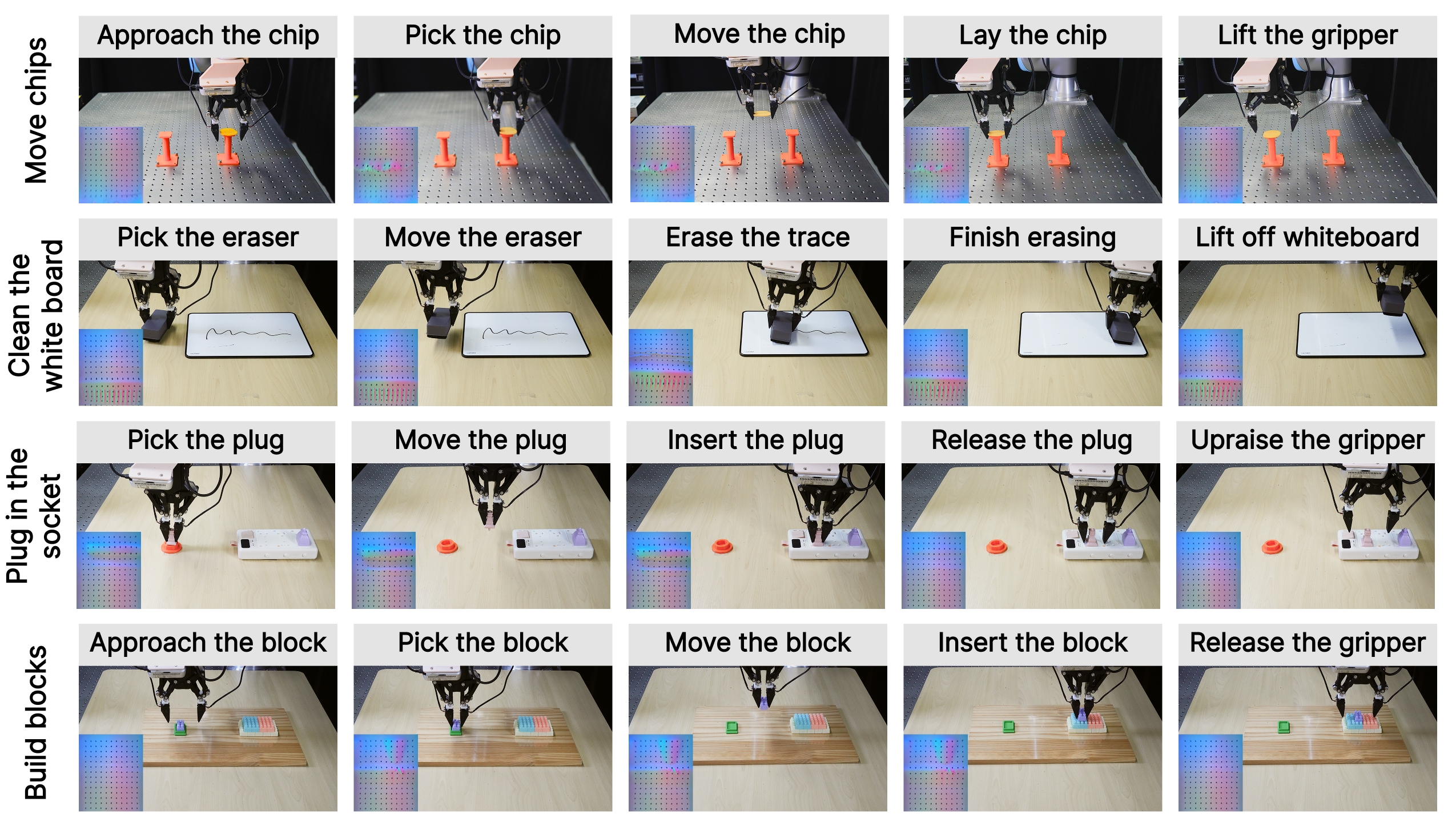}
    \caption{\textbf{Evaluation tasks.} Key stages of four benchmarks. From top to bottom: \textit{Chip Transfer}, \textit{Plug Insertion}, \textit{Whiteboard Wiping}, and \textit{Block Assembly}. Each row illustrates a typical execution sequence, with fingertip visuotactile observations visualized in the corner to reflect local deformations and interaction-state changes during contact.}
    \label{fig:tasks}
\end{figure*}

\section{EXPERIMENTS}
\label{sec:exp}

This section presents a systematic evaluation of the M\textsuperscript{2}-ResiPolicy framework, designed to address the competing multi-timescale requirements of contact-rich manipulation. Our experiments are structured to validate the three core contributions: the slow--fast hierarchical architecture, the tactile-force-adaptive fusion, and the impedance-controlled execution. We specifically address the following Research Questions (RQs):

\begin{itemize}
    \item \textbf{RQ1 (Perception Effectiveness):} Can the tactile-force-adaptive weighting mechanism effectively regulate the contribution of multimodal information across different contact phases (e.g., free space versus contact)?
    \item \textbf{RQ2 (Residual Correction):} Does the high-frequency \textit{GRU-Based Micro-Residual Corrector (MRC)} significantly outperform single-frequency baselines in compensating for contact dynamics and accumulated drift?
    \item \textbf{RQ3 (Interaction Safety):} Does integrating \textit{Position-Based Hybrid Impedance Control} provide superior safety and stability compared to rigid position control in constrained environments?
\end{itemize}

\subsection{Experimental Setup}
\label{subsec:setup}

\subsubsection{\textbf{System and Data Collection}}
As detailed in Sec.~\ref{sec:method}, our experimental platform comprises a UR7e manipulator teleoperated via a GELLO device. Perception is provided by two RealSense D435 cameras (global and wrist views) and two fingertip-mounted XENSE-G1 visuotactile sensors. To support our cross-frequency architecture, we record multimodal observations and master actions at 10~Hz, while synchronously logging high-fidelity TCP wrench data at 60~Hz for residual learning.

\subsubsection{\textbf{Task Definitions}}
To comprehensively evaluate policy performance under varying contact conditions, we select two categories of representative tasks, as visualized in Fig.~\ref{fig:tasks}:
\begin{itemize}
    \item \textbf{Category I: Force-Sensitive Manipulation.} Includes \textit{Fragile Chip Transfer} and \textit{Surface Wiping}. The primary challenge here is precise force regulation to prevent object damage or excessive contact pressure.
    \item \textbf{Category II: Geometry-Constrained Assembly.} Includes \textit{Dual-Peg Insertion} and \textit{Block Assembly}. These tasks involve millimeter-level tolerances and severe visual occlusion, requiring agile local correction to resolve geometric conflicts.
\end{itemize}

\subsubsection{\textbf{Baselines}}
\begin{figure*}[t]
    \centering
    \includegraphics[width=0.95\linewidth]{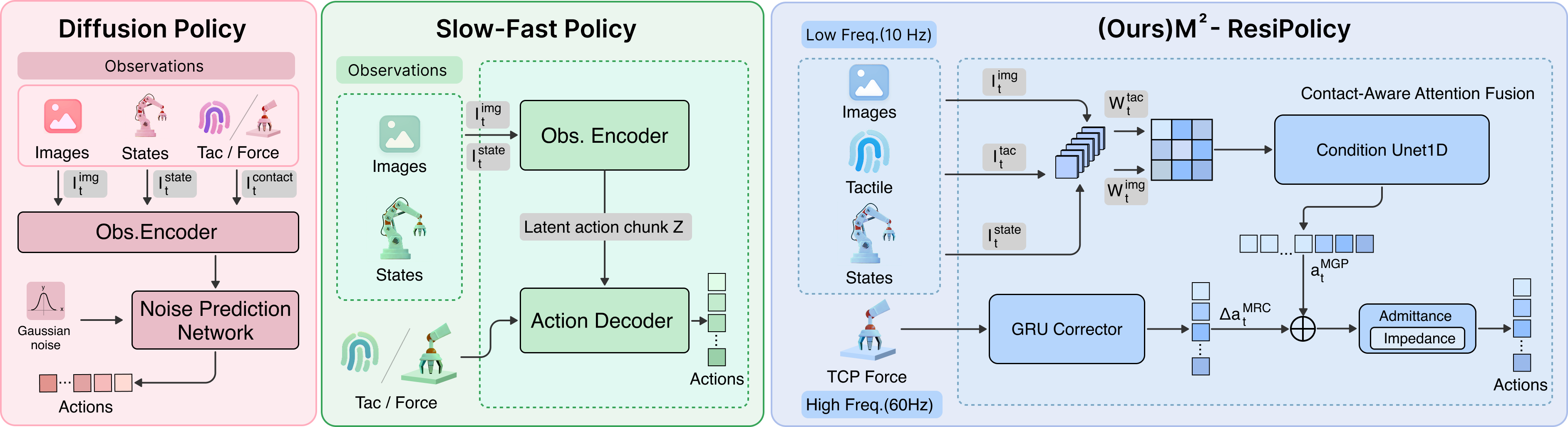}
    \caption{\textbf{Baseline Architectures.} We compare our hierarchical MGP+MRC framework against the open-loop Diffusion Policy (DP) and the end-to-end Reactive Diffusion Policy (RDP).}
    \label{fig:model_comparison}
\end{figure*}

We compare our method against two primary baselines (architectures shown in Fig.~\ref{fig:model_comparison}): (i) Diffusion Policy (DP), a standard diffusion-based imitation learning baseline without explicit high-frequency reactive correction; and (ii) Reactive Diffusion Policy (RDP), a representative slow--fast reactive policy baseline for contact-rich manipulation. To answer RQ2 and RQ3, we further evaluate three ablation variants: Ours (w/o Adaptive Fusion), Ours (w/o MRC), and Ours (Rigid Control).

\subsubsection{\textbf{Evaluation Protocols}}
Each task is evaluated over 30 independent trials per method. For methods with the impedance-based execution layer, the stiffness and damping parameters are tuned per task and kept fixed across all trials. In addition to binary success rates, we analyze failure modes and force profiles to provide a mechanistic understanding of policy behavior.

\subsection{Main Results}
\label{subsec:main_results}

\begin{figure}[!ht]
    \centering
    \includegraphics[width=0.9\linewidth]{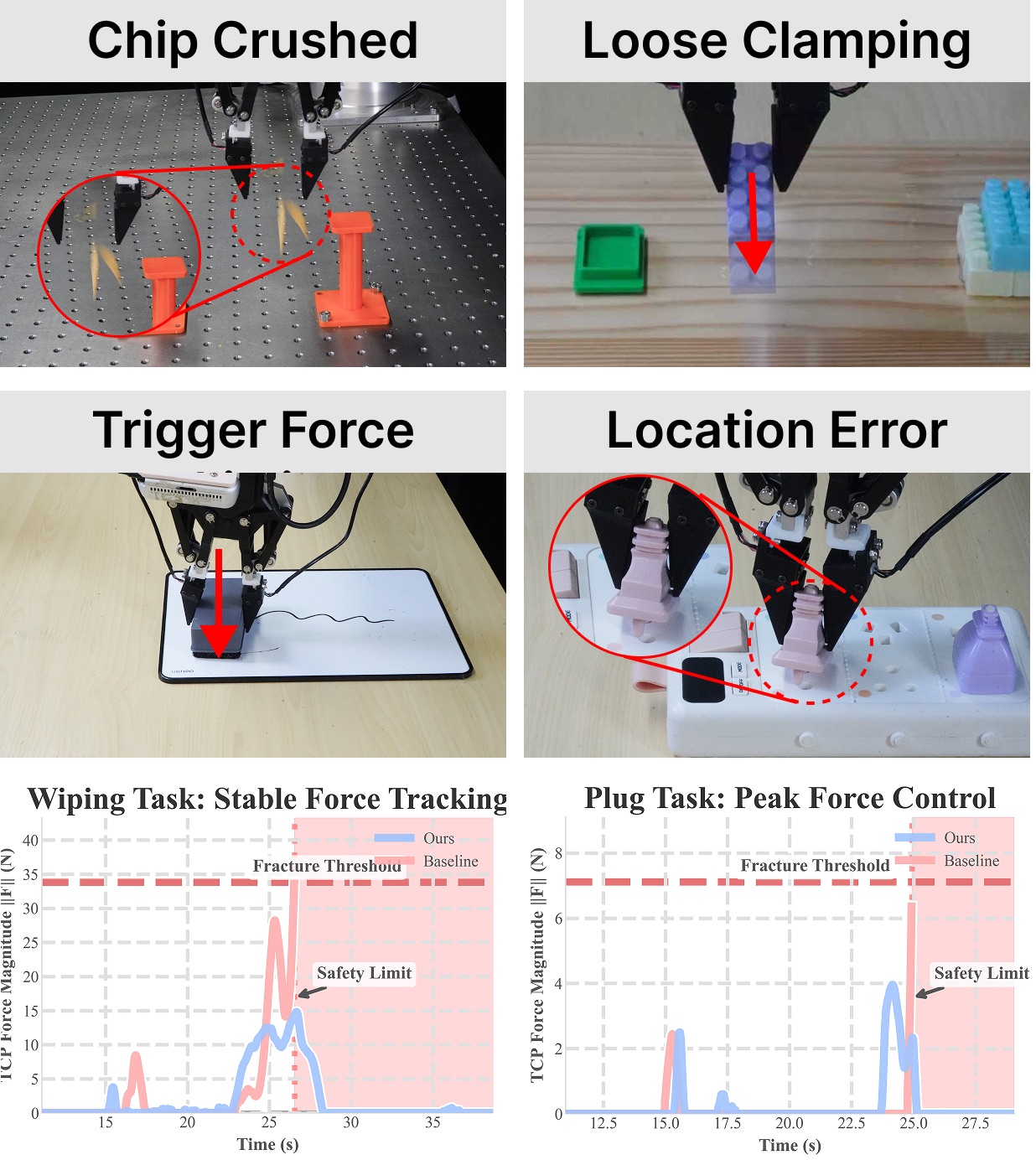}
    
    \caption{\textbf{Failure Analysis.} 
\textbf{(Top)} Baseline failures: \textit{Perception} errors cause crushing/slippage; \textit{Control} rigidity leads to force stops/jamming. 
\textbf{(Bottom)} Force profiles contrast unsafe baseline spikes (Red) with our stable regulation (Blue).}
    \label{fig:fail_analysis}
\end{figure}

We summarize the quantitative success rates across all tasks in Table~\ref{tab:main_results} and analyze the underlying failure mechanisms using the qualitative visualization in Fig.~\ref{fig:fail_analysis}. Based on these results, we address the three core research questions.

\begin{table*}[!ht]
    \centering
    \caption{\textbf{Policy Performance Across Task-Specific Execution Stages (Success Rate)}}
    \label{tab:main_results}
    \setlength{\tabcolsep}{2.5pt} 
    \footnotesize
    \resizebox{\textwidth}{!}{%
        \begin{tabular}{l|ccc|ccc|ccc|ccc|c}
            \toprule
            \multirow{2}{*}{\textbf{Method}} & \multicolumn{3}{c|}{\textbf{Chip Transfer}} & \multicolumn{3}{c|}{\textbf{Surface Wiping}} & \multicolumn{3}{c|}{\textbf{Plug Insertion}} & \multicolumn{3}{c|}{\textbf{Block Assembly}} & \textbf{Score} \\ 
            \cmidrule(lr){2-4} \cmidrule(lr){5-7} \cmidrule(lr){8-10} \cmidrule(lr){11-13} \cmidrule(lr){14-14}
             & Pick & Move & Place & Pick & Contact & Wipe & Grasp & Align & Insert & Grasp & Align & Stack & \textbf{Avg.} \\ \midrule
             
            DP (Baseline) & 43\% & 40\% & 37\% & 80\% & 67\% & 63\% & 70\% & 63\% & 43\% & 57\% & 53\% & 43\% & 0.55 \\
            DP w. tactile & 63\% & 57\% & 53\% & 83\% & 73\% & 70\% & 77\% & 70\% & 53\% & 67\% & 60\% & 50\% & 0.65 \\ \midrule
             
            RDP (Tactile) & 90\% & 87\% & 83\% & \textbf{97\%} & 93\% & 90\% & 90\% & 87\% & 73\% & 93\% & 87\% & 77\% & 0.87 \\
            RDP (Force) & 70\% & 63\% & 60\% & \textbf{97\%} & 93\% & 93\% & 87\% & 87\% & \textbf{80\%} & 87\% & 83\% & 80\% & 0.82 \\ \midrule
             
            \textbf{M\textsuperscript{2}-ResiPolicy} & \textbf{100\%} & \textbf{97\%} & \textbf{93\%} & \textbf{97\%} & \textbf{97\%} & \textbf{97\%} & \textbf{93\%} & \textbf{93\%} & \textbf{80\%} & \textbf{97\%} & \textbf{93\%} & \textbf{87\%} & \textbf{0.94} \\ \bottomrule
        \end{tabular}%
    }
\end{table*}

\noindent\textbf{Addressing RQ1 (Perception Effectiveness):} \\
Experimental results show that the \textit{Tactile-Force-Adaptive Weighting} mechanism plays a key role in regulating multimodal feature contributions across contact phases. As shown in Table~\ref{tab:main_results}, the baseline methods exhibit clear performance degradation during the grasp-related stages of challenging tasks. As qualitatively illustrated in Fig.~\ref{fig:fail_analysis} (Bottom, Row 1), the open-loop baseline fails to adapt to varying contact conditions, resulting in crushing of fragile chips due to excessive closure (Left) and slippage of small objects due to insufficient contact regulation (Right). In contrast, M\textsuperscript{2}-ResiPolicy maintains consistently high grasp success rates across these settings. By adaptively increasing the contribution of tactile cues upon contact, our method balances the sensitivity required to avoid over-compression with the stability needed to prevent slip.

\noindent\textbf{Addressing RQ2 (Residual Correction):} \\
Results on precision-critical tasks demonstrate the utility of the \textit{Micro-Residual Corrector (MRC)}. As shown in Table~\ref{tab:main_results}, single-frequency baselines exhibit a clear performance drop during the execution stages of \textit{Insertion} and \textit{Assembly}. Without high-frequency residual correction, these methods struggle to compensate for millisecond-scale contact dynamics, leading to the \textbf{geometric jamming} shown in Fig.~\ref{fig:fail_analysis} (Bottom, Row 2, Right). In contrast, with the 60~Hz GRU-based residual module, M\textsuperscript{2}-ResiPolicy maintains higher execution success rates, indicating that the proposed slow--fast architecture effectively mitigates accumulated drift and improves robustness under contact-rich execution.

\noindent\textbf{Addressing RQ3 (Interaction Safety):} \\
Results on constrained contact tasks demonstrate that the proposed force-mixed PBIC execution layer acts as a reliable safety buffer for physical interaction. As shown in Fig.~\ref{fig:fail_analysis}, rigid control leads to unsafe behaviors in both qualitative failures and force profiles. In the \textit{Wiping} task, where sustained contact is required, high position stiffness can produce abnormally large normal-force peaks and trigger a \textbf{Protective Stop} (Bottom, Left). In the \textit{Plug Insertion} task, the baseline exhibits a sharp force spike during insertion, indicating poor tolerance to contact misalignment and constrained interaction (Bottom, Right). In contrast, M\textsuperscript{2}-ResiPolicy leverages compliant execution to keep contact forces within safe operating limits, thereby improving task continuity and interaction stability.

\subsection{Ablation Studies}
\label{subsec:ablation}

To quantify the individual contributions of the core components in M\textsuperscript{2}-ResiPolicy, we evaluate three ablation variants. Table~\ref{tab:ablation} details the success rates in the most challenging phases of each task, illustrating the performance changes caused by removing specific modules.
\begin{table}[t]
    \centering
    \caption{\textbf{Ablation Analysis: Success Rate Drops}}
    \label{tab:ablation}
    \setlength{\tabcolsep}{3.5pt} 
    \footnotesize 
    \renewcommand{\arraystretch}{1.3} 
    
    \begin{tabular}{l|cccc} 
        \toprule
        \textbf{Variants} & \makecell{\textbf{Chip}\\\textbf{Transfer}} & \makecell{\textbf{Surface}\\\textbf{Wiping}} & \makecell{\textbf{Plug}\\\textbf{Insertion}} & \makecell{\textbf{Block}\\\textbf{Assembly}} \\ 
        \midrule
        
        \textbf{Full Model (Ours)} & \textbf{93\%} & \textbf{97\%} & \textbf{80\%} & \textbf{87\%} \\ 
        \midrule
        
        w/o Adaptive Fusion & 80\% \tiny($\downarrow$13) & 93\% \tiny($\downarrow$04) & 73\% \tiny($\downarrow$07) & 77\% \tiny($\downarrow$10) \\
        
        w/o MRC (Residual) & 90\% \tiny($\downarrow$03) & 93\% \tiny($\downarrow$04) & 67\% \tiny($\downarrow$13) & 70\% \tiny($\downarrow$17) \\
        
        w/o Impedance (Rigid) & 87\% \tiny($\downarrow$06) & 90\% \tiny($\downarrow$07) & 70\% \tiny($\downarrow$10) & 77\% \tiny($\downarrow$10) \\ 
        \bottomrule
    \end{tabular}
\end{table}

\subsubsection{\textbf{Impact of Tactile-Force-Adaptive Fusion}}
As shown in Table~\ref{tab:ablation}, removing the adaptive weighting mechanism (\textit{w/o Adaptive Fusion}) leads to a consistent performance drop across all tasks, with the largest degradation observed in \textit{Chip Transfer} and \textit{Block Assembly}. In this variant, we remove the contact-dependent confidence gating while keeping the same visuotactile fusion architecture. Specifically, the dynamic mask is replaced by a constant all-one mask, such that tactile features always participate in fusion with a fixed weight regardless of contact state.

This result highlights the importance of phase-adaptive modality weighting. Without contact-aware gating, the policy cannot appropriately regulate the contribution of tactile cues across interaction phases, making it more difficult to balance competing sensitivity requirements. Consequently, the policy may apply excessive force when handling fragile objects or fail to detect incipient slip when manipulating small objects with limited contact margins. The consistent degradation across tasks suggests that adaptive fusion is critical for maintaining grasp stability and handling objects with diverse physical properties.

\subsubsection{\textbf{Impact of GRU Micro-Residual Correction}}
Removing the high-frequency residual module (\textit{w/o MRC}) causes the largest performance drop in \textit{Plug Insertion} and \textit{Block Assembly}, with success rates decreasing by 13\% and 17\%, respectively. This result suggests that the 10~Hz master policy alone is insufficient for geometry-constrained tasks with millimeter-level tolerances. Although the master branch can provide coarse alignment, without high-frequency force-based correction the system becomes less effective at compensating for local contact events during the final execution stage. These results highlight the role of the fast residual stream in improving precision-critical contact execution.

\subsubsection{\textbf{Impact of Impedance Control}}
Replacing the compliant execution layer with rigid position servoing (\textit{w/o Impedance}) causes a clear performance drop, particularly in \textit{Plug Insertion}, \textit{Block Assembly}, and \textit{Surface Wiping}. This result indicates that rigid position control has limited tolerance to contact misalignment and force spikes in sustained or constrained interaction. As supported by the failure analysis in Fig.~\ref{fig:fail_analysis}, this variant is more likely to trigger unsafe contact responses, including large force peaks and protective stops. The performance gap confirms that physical compliance is essential for robust contact-rich manipulation.

\subsection{Generalization Analysis}
\label{subsec:generalization}

\begin{figure}[!ht]
    \centering
    \includegraphics[width=0.9\linewidth]{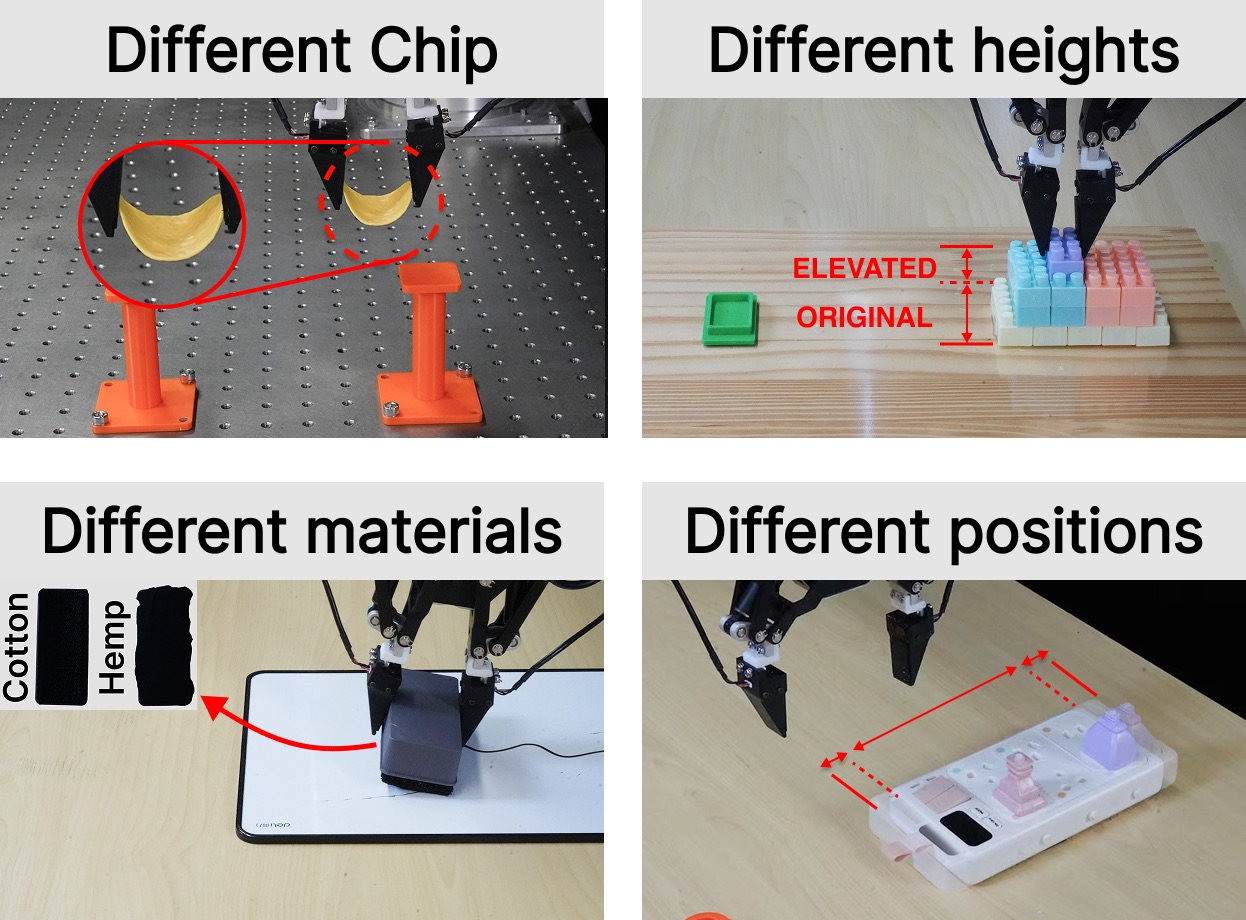}
    \caption{\textbf{Zero-Shot Generalization Scenarios.}}
    \label{fig:robustness_visual}
\end{figure}

To evaluate the robustness of M\textsuperscript{2}-ResiPolicy to out-of-distribution conditions, we introduce physical and geometric perturbations that are absent from the training data. The zero-shot generalization capability of the policy is assessed under four challenging scenarios, as illustrated in Fig.~\ref{fig:robustness_visual}.

\subsubsection{\textbf{Adaptation to Unseen Object Variations}}
We first evaluate robustness to unseen variations in object properties.
\begin{itemize}
    \item \textbf{Stiffness and Shape Variation:} For chips with different stiffness and shapes, rigid baselines tend to crush the object due to altered contact-force gradients and load distribution. In contrast, M\textsuperscript{2}-ResiPolicy benefits from confidence-gated visuotactile fusion in the master policy, which adjusts perceptual emphasis after contact and generates more appropriate grasp actions, thereby reducing the risk of breakage.
    \item \textbf{Wiping Material Variation:} When wiping with different materials, conventional control often fails to accommodate changes in tangential resistance and contact conditions, leading to incomplete wiping or visible residue. By contrast, M\textsuperscript{2}-ResiPolicy leverages compliant execution to maintain more stable contact and wiping performance under varying material conditions.
\end{itemize}

\subsubsection{\textbf{Robustness to Geometric Uncertainty}}
We further introduce positioning errors to simulate calibration drift.
\begin{itemize}
    \item \textbf{Vertical Error (+0.8cm):} While rigid policies often trigger protective stops under premature contact, the hybrid impedance layer regulates downward motion by converting position error into a safe normal force.
    \item \textbf{Lateral Offset (1cm):} Under lateral misalignment, the insertion trajectory predicted by the master policy deviates from the true socket position. After initial contact, asymmetric contact cues are captured by the confidence-gated visuotactile fusion, which triggers the residual policy to generate local corrections ($\Delta a$) and steer the peg back toward the correct insertion path.
\end{itemize}

\section{CONCLUSIONS}
We presented \textit{M}$^{2}$-\textit{ResiPolicy}, a Master--Micro residual control framework for contact-rich and fine-grained manipulation. The proposed framework combines confidence-gated visuotactile action generation in the master policy, high-rate residual correction from TCP wrench feedback, and a force-mixed PBIC execution layer for compliant interaction. In this way, it unifies long-horizon motion consistency with rapid contact adaptation, and integrates a force-mixed PBIC execution layer to ensure inherently safe and stable interaction under contact. Experiments on multiple challenging tasks demonstrate that \textit{M}$^{2}$-\textit{ResiPolicy} consistently outperforms strong diffusion-based baselines in both task success rate and force-regulation stability, validating the effectiveness of coupling multi-timescale policy learning with compliant execution.

Despite these encouraging results, several limitations remain. In particular, the low-level force-control parameters are still tuned empirically and do not adapt online to changing interaction conditions. Future work will explore adaptive low-level control mechanisms, tighter integration of higher-rate tactile and force feedback, and more unified optimization between policy generation and compliant execution to further improve robustness and generalization.

\addtolength{\textheight}{-12cm}   









\bibliographystyle{ieeetr}
\bibliography{references}

@article{liu2025vitamin,
  title={Vitamin: Learning contact-rich tasks through robot-free visuo-tactile manipulation interface},
  author={Liu, Fangchen and Li, Chuanyu and Qin, Yihua and Xu, Jing and Abbeel, Pieter and Chen, Rui},
  journal={arXiv preprint arXiv:2504.06156},
  year={2025}
}

@inproceedings{li2025adaptive,
  title={Adaptive visuo-tactile fusion with predictive force attention for dexterous manipulation},
  author={Li, Jinzhou and Wu, Tianhao and Zhang, Jiyao and Chen, Zeyuan and Jin, Haotian and Wu, Mingdong and Shen, Yujun and Yang, Yaodong and Dong, Hao},
  booktitle={2025 IEEE/RSJ International Conference on Intelligent Robots and Systems (IROS)},
  pages={3232--3239},
  year={2025},
  organization={IEEE}
}

@article{xue2025reactive,
  title={Reactive diffusion policy: Slow-fast visual-tactile policy learning for contact-rich manipulation},
  author={Xue, Han and Ren, Jieji and Chen, Wendi and Zhang, Gu and Fang, Yuan and Gu, Guoying and Xu, Huazhe and Lu, Cewu},
  journal={arXiv preprint arXiv:2503.02881},
  year={2025}
}

@inproceedings{liu2025forcemimic,
  title={Forcemimic: Force-centric imitation learning with force-motion capture system for contact-rich manipulation},
  author={Liu, Wenhai and Wang, Junbo and Wang, Yiming and Wang, Weiming and Lu, Cewu},
  booktitle={2025 IEEE International Conference on Robotics and Automation (ICRA)},
  pages={1105--1112},
  year={2025},
  organization={IEEE}
}

@article{chen2025dexforce,
  title={Dexforce: Extracting force-informed actions from kinesthetic demonstrations for dexterous manipulation},
  author={Chen, Claire and Yu, Zhongchun and Choi, Hojung and Cutkosky, Mark and Bohg, Jeannette},
  journal={IEEE Robotics and Automation Letters},
  year={2025},
  publisher={IEEE}
}

@article{xie2025universal,
  title={Universal Visuo-Tactile Video Understanding for Embodied Interaction},
  author={Xie, Yifan and Li, Mingyang and Li, Shoujie and Li, Xingting and Chen, Guangyu and Ma, Fei and Yu, Fei Richard and Ding, Wenbo},
  journal={arXiv preprint arXiv:2505.22566},
  year={2025}
}

@article{heng2025vitacformer,
  title={ViTacFormer: Learning Cross-Modal Representation for Visuo-Tactile Dexterous Manipulation},
  author={Heng, Liang and Geng, Haoran and Zhang, Kaifeng and Abbeel, Pieter and Malik, Jitendra},
  journal={arXiv preprint arXiv:2506.15953},
  year={2025}
}

@inproceedings{zhao2025cot,
  title={Cot-vla: Visual chain-of-thought reasoning for vision-language-action models},
  author={Zhao, Qingqing and Lu, Yao and Kim, Moo Jin and Fu, Zipeng and Zhang, Zhuoyang and Wu, Yecheng and Li, Zhaoshuo and Ma, Qianli and Han, Song and Finn, Chelsea and others},
  booktitle={Proceedings of the Computer Vision and Pattern Recognition Conference},
  pages={1702--1713},
  year={2025}
}

@inproceedings{jones2025beyond,
  title={Beyond sight: Finetuning generalist robot policies with heterogeneous sensors via language grounding},
  author={Jones, Joshua and Mees, Oier and Sferrazza, Carmelo and Stachowicz, Kyle and Abbeel, Pieter and Levine, Sergey},
  booktitle={2025 IEEE International Conference on Robotics and Automation (ICRA)},
  pages={5961--5968},
  year={2025},
  organization={IEEE}
}

@article{yu2025forcevla,
  title={ForceVLA: Enhancing VLA Models with a Force-aware MoE for Contact-rich Manipulation},
  author={Yu, Jiawen and Liu, Hairuo and Yu, Qiaojun and Ren, Jieji and Hao, Ce and Ding, Haitong and Huang, Guangyu and Huang, Guofan and Song, Yan and Cai, Panpan and others},
  journal={arXiv preprint arXiv:2505.22159},
  year={2025}
}

@article{jiang2025gelfusion,
  title={GelFusion: Enhancing Robotic Manipulation under Visual Constraints via Visuotactile Fusion},
  author={Jiang, Shulong and Zhao, Shiqi and Fan, Yuxuan and Yin, Peng},
  journal={arXiv preprint arXiv:2505.07455},
  year={2025}
}

@article{liu2025factr,
  title={Factr: Force-attending curriculum training for contact-rich policy learning},
  author={Liu, Jason Jingzhou and Li, Yulong and Shaw, Kenneth and Tao, Tony and Salakhutdinov, Ruslan and Pathak, Deepak},
  journal={arXiv preprint arXiv:2502.17432},
  year={2025}
}

@article{he2025foar,
  title={FoAR: Force-Aware Reactive Policy for Contact-Rich Robotic Manipulation},
  author={He, Zihao and Fang, Hongjie and Chen, Jingjing and Fang, Hao-Shu and Lu, Cewu},
  journal={IEEE Robotics and Automation Letters},
  year={2025},
  publisher={IEEE}
}

@inproceedings{wu2025tacdiffusion,
  title={Tacdiffusion: Force-domain diffusion policy for precise tactile manipulation},
  author={Wu, Yansong and Chen, Zongxie and Wu, Fan and Chen, Lingyun and Zhang, Liding and Bing, Zhenshan and Swikir, Abdalla and Haddadin, Sami and Knoll, Alois},
  booktitle={2025 IEEE International Conference on Robotics and Automation (ICRA)},
  pages={11831--11837},
  year={2025},
  organization={IEEE}
}

@article{kang2025robotic,
  title={Robotic compliant object prying using diffusion policy guided by vision and force observations},
  author={Kang, Jeon Ho and Joshi, Sagar and Huang, Ruopeng and Gupta, Satyandra K},
  journal={IEEE Robotics and Automation Letters},
  year={2025},
  publisher={IEEE}
}

@inproceedings{hou2025adaptive,
  title={Adaptive compliance policy: Learning approximate compliance for diffusion guided control},
  author={Hou, Yifan and Liu, Zeyi and Chi, Cheng and Cousineau, Eric and Kuppuswamy, Naveen and Feng, Siyuan and Burchfiel, Benjamin and Song, Shuran},
  booktitle={2025 IEEE International Conference on Robotics and Automation (ICRA)},
  pages={4829--4836},
  year={2025},
  organization={IEEE}
}

@article{yan2025maniflow,
  title={Maniflow: A general robot manipulation policy via consistency flow training},
  author={Yan, Ge and Zhu, Jiyue and Deng, Yuquan and Yang, Shiqi and Qiu, Ri-Zhao and Cheng, Xuxin and Memmel, Marius and Krishna, Ranjay and Goyal, Ankit and Wang, Xiaolong and others},
  journal={arXiv preprint arXiv:2509.01819},
  year={2025}
}

@inproceedings{hansen2022visuotactile_rl,
  title={Visuotactile-rl: Learning multimodal manipulation policies with deep reinforcement learning},
  author={Hansen, Johanna and Hogan, Francois and Rivkin, Dmitriy and Meger, David and Jenkin, Michael and Dudek, Gregory},
  booktitle={2022 International Conference on Robotics and Automation (ICRA)},
  pages={8298--8304},
  year={2022},
  organization={IEEE}
}

@article{zhao2025touch_begins,
  title={Touch begins where vision ends: Generalizable policies for contact-rich manipulation},
  author={Zhao, Zifan and Haldar, Siddhant and Cui, Jinda and Pinto, Lerrel and Bhirangi, Raunaq},
  journal={arXiv preprint arXiv:2506.13762},
  year={2025}
}

@article{huang2025vt_refine,
  title={Vt-refine: Learning bimanual assembly with visuo-tactile feedback via simulation fine-tuning},
  author={Huang, Binghao and Xu, Jie and Akinola, Iretiayo and Yang, Wei and Sundaralingam, Balakumar and O'Flaherty, Rowland and Fox, Dieter and Wang, Xiaolong and Mousavian, Arsalan and Chao, Yu-Wei and others},
  journal={arXiv preprint arXiv:2510.14930},
  year={2025}
}

@article{helmut2025tactile_conditioned_diffusion,
  title={Tactile-Conditioned Diffusion Policy for Force-Aware Robotic Manipulation},
  author={Helmut, Erik and Funk, Niklas and Schneider, Tim and de Farias, Cristiana and Peters, Jan},
  journal={arXiv preprint arXiv:2510.13324},
  year={2025}
}

@article{yang2022touch_and_go,
  title={Touch and go: Learning from human-collected vision and touch},
  author={Yang, Fengyu and Ma, Chenyang and Zhang, Jiacheng and Zhu, Jing and Yuan, Wenzhen and Owens, Andrew},
  journal={arXiv preprint arXiv:2211.12498},
  year={2022}
}

@inproceedings{wu2024gello,
  title={Gello: A general, low-cost, and intuitive teleoperation framework for robot manipulators},
  author={Wu, Philipp and Shentu, Yide and Yi, Zhongke and Lin, Xingyu and Abbeel, Pieter},
  booktitle={2024 IEEE/RSJ International Conference on Intelligent Robots and Systems (IROS)},
  pages={12156--12163},
  year={2024},
  organization={IEEE}
}

@article{vaswani2017attention,
  title={Attention is all you need},
  author={Vaswani, Ashish and Shazeer, Noam and Parmar, Niki and Uszkoreit, Jakob and Jones, Llion and Gomez, Aidan N and Kaiser, {\L}ukasz and Polosukhin, Illia},
  journal={Advances in neural information processing systems},
  volume={30},
  year={2017}
}

@article{chi2025diffusion,
  title={Diffusion policy: Visuomotor policy learning via action diffusion},
  author={Chi, Cheng and Xu, Zhenjia and Feng, Siyuan and Cousineau, Eric and Du, Yilun and Burchfiel, Benjamin and Tedrake, Russ and Song, Shuran},
  journal={The International Journal of Robotics Research},
  volume={44},
  number={10-11},
  pages={1684--1704},
  year={2025},
  publisher={Sage Publications Sage UK: London, England}
}

@article{ge2025filic,
  title={FILIC: Dual-Loop Force-Guided Imitation Learning with Impedance Torque Control for Contact-Rich Manipulation Tasks},
  author={Ge, Haizhou and Jia, Yufei and Li, Zheng and Li, Yue and Chen, Zhixing and Huang, Ruqi and Zhou, Guyue},
  journal={arXiv preprint arXiv:2509.17053},
  year={2025}
}

\end{document}